\documentclass[preprint,3p,times,twocolumn]{elsarticle}

\usepackage{amssymb}
\usepackage{amsmath}
\usepackage[colorlinks, citecolor=blue]{hyperref}

\journal{Information Fusion}
\begin{document}

\begin{frontmatter}



\title{Conti-Fuse: A Novel Continuous Decomposition-based Fusion Framework for Infrared and Visible Images}


\author[label1]{Hui Li\corref{cor1}}
\author[label1]{Haolong Ma}
\author[label1]{Chunyang Cheng}
\author[label2]{Zhongwei Shen}
\author[label1]{Xiaoning Song}
\author[label1]{Xiao-Jun Wu}

\affiliation[label1]{organization={School of Artificial Intelligence and Computer Science, Jiangnan University},
            addressline={214122}, 
            city={Wuxi},
            country={China}}
\affiliation[label2]{organization={School of Electronic \& Information Engineering, Suzhou University of Science and Technology},
            addressline={215009}, 
            city={Suzhou},
            country={China}}

\cortext[cor1]{Corresponding author email: lihui.cv@jiangnan.edu.cn}

\begin{abstract}
For better explore the relations of inter-modal and inner-modal, even in deep learning fusion framework, the concept of decomposition plays a crucial role. However, the previous decomposition strategies (base \& detail or low-frequency \& high-frequency) are too rough to present the common features and the unique features of source modalities, which leads to a decline in the quality of the fused images. The existing strategies treat these relations as a binary system, which may not be suitable for the complex generation task (e.g. image fusion).
To address this issue, a continuous decomposition-based fusion framework (Conti-Fuse) is proposed. Conti-Fuse treats the decomposition results as few samples along the feature variation trajectory of the source images, extending this concept to a more general state to achieve continuous decomposition. This novel continuous decomposition strategy enhances the representation of complementary information of inter-modal by increasing the number of decomposition samples, thus reducing the loss of critical information. To facilitate this process, the continuous decomposition module (CDM) is introduced to decompose the input into a series continuous components. The core module of CDM, State Transformer (ST), is utilized to efficiently capture the complementary information from source modalities. Furthermore, a novel decomposition loss function is also designed which ensures the smooth progression of the decomposition process while maintaining linear growth in time complexity with respect to the number of decomposition samples. Extensive experiments demonstrate that our proposed Conti-Fuse achieves superior performance compared to the state-of-the-art fusion methods.

\end{abstract}




\end{frontmatter}




\section{Introduction}
As a fundamental field of image processing, image fusion seeks to create informative and visually appealing images by extracting the most significant information from various source images \cite{xu2021stereo, yang2022sir, zhao2023metafusion, zhang2021image}. One of the notable challenges in image processing is Infrared and Visible Image Fusion (IVIF), which entails integrating complementary information from distinct modalities \cite{tang2023divfusion,liu2023sgfusion,li2021rfn}. In the IVIF task, the input comprises both infrared and visible images. Visible images are distinguished by their abundant texture information, which aligns more closely with human visual perception. However, they are susceptible to lighting variations, occlusion, and other factors, resulting in the loss of vital information. In contrast, infrared images excel in highlighting targets in extreme conditions (e.g., low light) by capturing thermal radiation but are prone to noise. Consequently, in IVIF tasks, the fused image must mitigate the shortcomings of both modalities to achieve superior visual quality\cite{hermessi2021multimodal}.

\begin{figure*}[tb]
  \centering
  \includegraphics[width=1\textwidth]{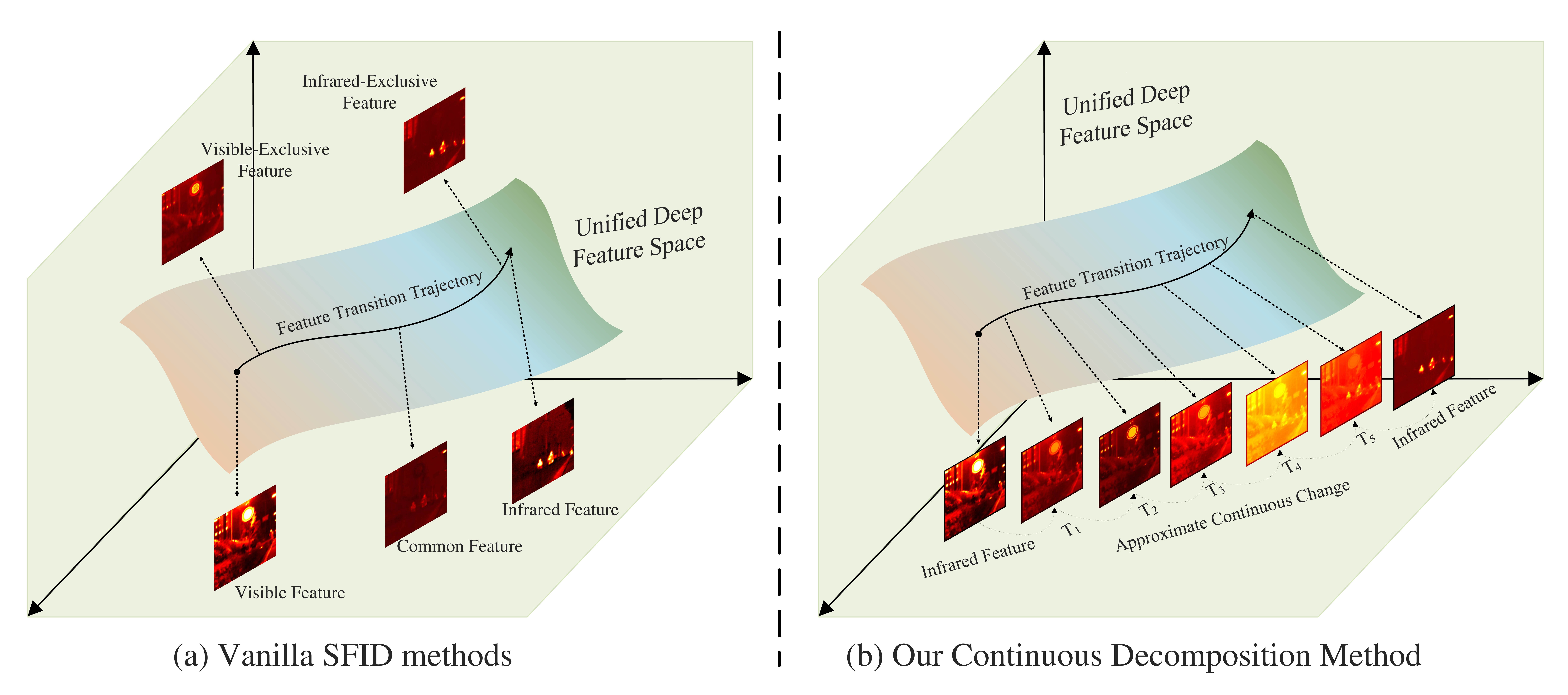}
  \caption{A schematic of the unified deep feature space between the common SFID(e.g. DeFusion \cite{liang2022fusion}) methods and our proposed continuous decomposition method.
  }
  \label{fig:intro}
\end{figure*}

Image decomposition, as a crucial technique, is frequently applied in image fusion tasks. Depending on the spatial domain in which the decomposition occurs, image decomposition methods can be broadly classified into the Shallow Feature space-based Image Decomposition Methods (SFID) \cite{lu2014novel, li2017pixel} and the Deep Feature space-based Image Decomposition Methods (DFID) \cite{liang2022fusion, li2023lrr, zhao2020didfuse}. In early traditional image fusion methods, SFIDs are the most prevalent. Among these, multi-scale transform (MST)-based decomposition methods \cite{pajares2004wavelet, shen2014exposure} are particularly popular. These methods employ manually design decomposition operations, such as discrete Fourier transforms, to decompose the original image into coefficients at multiple scales. Subsequently, fusion strategies and corresponding inverse transformations are applied to the coefficients to obtain the fused image. These methods achieve excellent results in early image fusion methods. However, since they only perform decomposition at the shallow feature space using manually designed strategies, these SFIDs lack adaptability to the source images, resulting in poor generalization capabilities.

To overcome the limitations of SFID, many approaches utilize deep neural networks to decompose at deeper feature space \cite{zhang2021sdnet, li2020mdlatlrr, tang2022mdedfusion}, representing source images. These methods typically consist of Encoders, Decomposition Modules, Fusion Modules, and Decoders. The Encoder is responsible for extracting basic features from the source images and mapping them into the Unified Deep Feature Space (UDFS) for representation. The UDFS refers to a unified feature space where two different input modalities are mapped. This enables the representation of information from different modalities within a unified space, laying the foundation for the subsequent image decomposition process. Subsequently, the Decomposition Module decomposes the features of the source images within this unified feature space into several decomposed features. Finally, these decomposed features are fused by the Fusion Module and mapped back to the pixel space by the Decoder to obtain the fused image. 

Taking DeFusion \cite{liang2022fusion} as an example, it uses convolutional networks as the Encoder to decompose the two source images into a shared feature and tow modality-specific features within a feature space. These three sets of decomposed features are then fused and reconstructed by a learnable convolutional network to produce the fused result. However, these methods roughly decompose the original images into few non-overlapping features, leading to the loss of some critical information from the source images (such as detailed information).

To address these shortcomings, we propose a continuous decomposition-based fusion framework, referred to as Conti-Fuse. As illustrated in Fig.\ref{fig:intro} (a), we consider the decomposed features of two source images as sample points along a continuous change trajectory in the unified deep feature space from one source image feature to another. Taking DeFusion \cite{liang2022fusion} as an example, its common features can be viewed as sample points in the middle of it's trajectory, while the two unique features can be approximated as sample points near the two ends of it's trajectory. Based on this concept, as depicted in Fig.\ref{fig:intro} (b), the proposed Conti-Fuse generalizes the decomposition in the unified deep feature space beyond the past simple decomposition method such as the common and unique features in DeFusion \cite{liang2022fusion}. By performing multiple decompositions along the continuous transition trajectory in the feature space, we obtain richer and more diverse decomposed features, referred to as transition states, which more finely represent the critical information of the two source images.

Inspired by MST-based methods \cite{nunez1999multiresolution,pohl1998review}, we apply our continuous decomposition method to a multi-scale framework and propose the Continuous Decomposition Module (CDM) to decompose features and obtain transition states. Additionally, to fully capture the complementary information between transition states, we introduce the State Transformer to enhance the complementarity between transition states. Finally, to guide the entire decomposition process, we design a novel continuous decomposition loss function and the corresponding computational strategy, termed as Support Decomposition Strategy (SDS). SDS employs the Monte Carlo method to perform random sampling of the continuous decomposition loss, approximating the true decomposition loss with sufficient loss samples, thereby reducing the computational complexity of the continuous decomposition loss from quadratic to linear, which accelerates the training speed of the model.

The contributions of our work are summarized as follows: 

\begin{itemize}
\item A novel decomposition strategy is introduced, which achieves enriched decomposition features by densely sampling along the variation trajectories of deep features across the two modalities. This method effectively reduces the loss of crucial information in fused images.

\item An efficient decomposition loss is designed to facilitate continuous decomposition. By leveraging the Monte Carlo method, this loss function accelerates computation, thereby enhancing the scalability of the proposed approach.

\item Extensive qualitative and quantitative experiments were conducted, demonstrating excellent performance of our approach compared to other state-of-the-art fusion methods.
\end{itemize}

\section{Related Work}
\subsection{Shallow Feature Space-based Image Decomposition (SFID) Methods}
SFID methods are commonly used in image fusion \cite{panda2024weight, panda2020edge, panda2020pixel, panda2020pixel}. Among these methods, multi-scale transformation-based decomposition techniques are particularly popular \cite{pajares2004wavelet, shen2014exposure, panda2024bayesian}. Typically, multi-scale transformation methods involve three steps \cite{tang2023datfuse}: (1) Decomposing the original image into coefficients at various scales using a specific transformation, (2) Integrating the coefficients of different modalities through carefully designed fusion rules, (3) Generating the fused image from the aggregated coefficients using the inverse transformation. 

For instance, Pajares et al. \cite{pajares2004wavelet} proposed a fusion model based on discrete wavelet transform decomposition. This model decomposed two source images into mult-scale coefficients using discrete wavelet transform, followed by manually designed merging strategies and inverse discrete wavelet transform to obtain the fused image. These models reliance on manually designed decomposition operations and fusion strategies lead to poor robustness of the models.

Compared to SFID methods, our method does not rely on manually designed decomposition operations and fusion strategies, providing better robustness.

\subsection{Deep Feature space-based Image Decomposition (DFID) Methods}
With the development of deep learning, many image fusion methods leverage the powerful representation capabilities of deep learning for image decomposition \cite{zhou2023perceptual,li2023infrared, panda2022integration, li2020fast}. These methods typically include an Encoder and a Decoder. The Encoder maps the two source images from the original pixel space to a unified deep feature space for representation. The Decoder maps the fused features, rich in deep semantic information, back to the original pixel space to obtain the fused image \cite{liang2022fusion}. The decomposition and fusion processes of features of source images are carried out in the unified deep feature space to ensure the model learns robust representations of them. 

This representation can be further uniformly described as follows \cite{zhang2023visible}: In a high-dimensional unified feature space, the decomposed features are considered as several sampling points on a feature transition trajectory between the source images. For example, in DeFusion \cite{liang2022fusion}, the three decomposed features (infrared unique features, common features, and visible unique features) are considered as three feature sampling points on the variation trajectory from infrared to visible features. Its recent successor, DeFusion++\cite{liang2024fusion}, follows the same decomposition strategy of common and unique features. However, existing DFID crudely decompose the source images into non-overlapping multiple features, which are sparse sampling points. This leads to insufficient representation of source images, losing much critical information.

Compared to existing DFID methods, our method offers a more general decomposition strategy with richer decomposed features, capable of preserving key information in the source images, thereby improving image quality.


\begin{figure*}[tb]
  \centering
  \includegraphics[width=1\textwidth,height=0.67\textwidth]{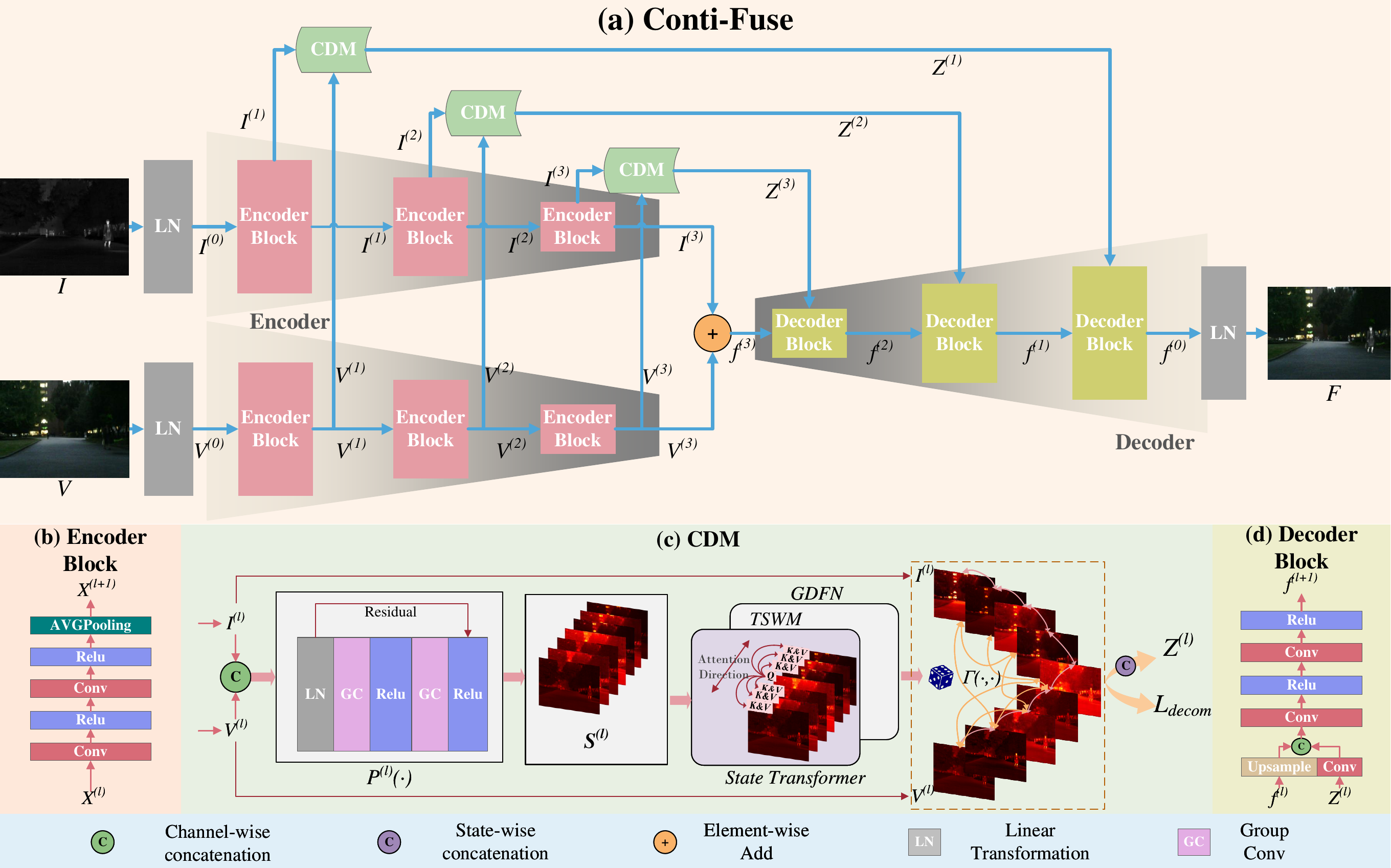}
  \caption{The architecture of Conti-Fuse. (a) The pipeline of proposed method. (b, c, d) The internal structure diagrams for Encoder Block, CDM and Decoder Block in the \( l \)-th layer, respectively. The input of Encoder Block ($X^{(l)}$) can be visible feature or infrared feature. 'Channel-wise concatenation' and 'State-wise concatenation' refer to concatenation along the channel and state dimensions of the tensors, respectively; 'Linear Transformation' refers to a 1 × 1 convolution, and 'Group Conv' refers to grouped convolution.
  }
  \label{fig:arch}
\end{figure*}

\begin{figure*}[tb]
  \centering
  \includegraphics[width=1\textwidth,height=0.31\textwidth]{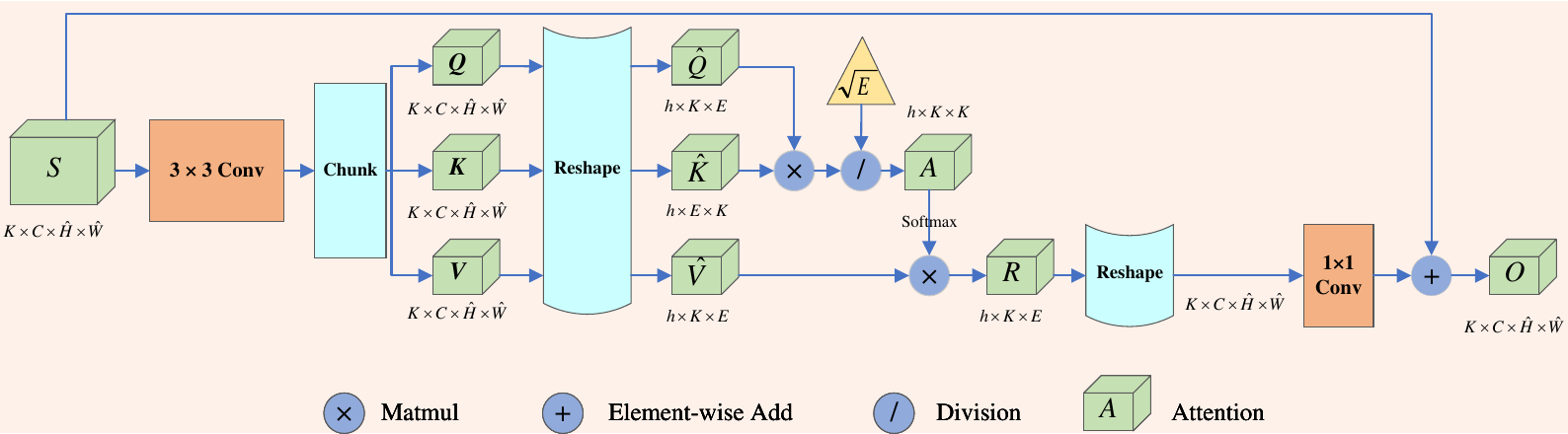}
  \caption{Illustration of Transition State Wise MHSA (TSWM).}
  \label{fig:st}
\end{figure*}

\section{Proposed Method}
In this section, firstly, the workflow of our proposed model and the detailed design of each module are introduced. Then, we provide the formulas for calculating the proposed loss function and explain the underlying design principles.

\subsection{Overview}
Conti-Fuse is mainly composed of three types of modules: Encoder, Decoder and The Continuous Decomposition Module (CDM). Encoder and Decoder are designed to extract shallow features from the source images and reconstructing the fused image, respectively. CDM is designed for interaction between two modalities and for generating transition states. In addition to the above three modules, there are two linear transformations (LN) at the input and output positions to adjust the number of feature channels. 

As shown in Fig.\ref{fig:arch} (a), the multi-scale structure is adopted in the proposed framework. The Encoder and Decoder have several Blocks and are marked $EN^{(l)}$ and $ DE^{(l)} (l\in \{1,2,\cdots,N\}, N=3$), respectively. For more general discussion, we set the number of transition states to $K$ and the number of layers to $N$ in this section.

\subsection{Encoder Block}
As shown in Fig.\ref{fig:arch} (b), the Encoder Block is composed of two convolutional layers with $3\times 3$ kernels, two ReLU activation functions, and one average pooling ($2\times 2$ ). These settings are shared for both infrared and visible branches. The Encoder Block is utilized to extract basic shallow features while mapping source images into a Unified Deep Feature Space (UDFS).

In this section, we introduce the following notation: $I, V\in \mathbb R^{H\times W}$ denote the input infrared image and visible image. Thus, the processing of input by the $l$-th layer Encoder Block can be expressed as follows,
\begin{equation}
\begin{gathered}
    I^{(l)}=EN^{(l)}(I^{(l - 1)}),V^{(l)}=EN^{(l)}(V^{(l - 1)})\\
    s.t. \quad l \in \{1,2,\cdots,N\}
\end{gathered}
\end{equation}
where $I^{(0)}$ and $V^{(0)}$ are obtained by the input linear transformation (LN, $1\times 1$ Conv) of the sources image $I$ and $V$, respectively.

\subsection{Continuous Decomposition Module}
The Continuous Decomposition Module (CDM) is proposed to decompose the input features into a continuous transition states, approximately. To do this, it is necessary to first extract information within each transition states, and then capture the complementary information from multiple modalities. 

As illustrated in Fig.\ref{fig:arch} (c), a linear transformation and several group convolutions are employed. The linear transformation are utilized to generate preliminary transition states. Group convolutions further extract finer information within each transition state. 

Concretely, $3\times3$ convolutional layer is employed in which the number of groups is set to $K$. Then, we obtain $K$ transition states denoted as $S\in \mathbb R^{K\times C\times \hat H \times \hat W}$, where $C$ represents the number of channels, and $\hat H \times \hat W$ represents the size of the feature maps of transition states. The sub-module in CDM is denoted as $P(\cdot)$  which is consisted by a linear transformation, two group convolutional layers and two ReLU activation functions.

\subsubsection{State Transformer}
To leverage the complementary information as much as possible, a novel feature extractor is designed which is named State Transformer, $ST(\cdot)$. The core module of ST is the \textbf{T}ransition \textbf{S}tate \textbf{W}ise \textbf{M}HSA (TSWM), which leverages the multi-head self-attention mechanism to capture complementary relationships between transition states.

As shown in Fig.\ref{fig:st}, the TSWM first generate $\mathbf Q$, $\mathbf K$, and $\mathbf V$ from $S$, which is accomplished through a linear transformation. Then, we reshape and split $\mathbf Q$, $\mathbf K$ and $\mathbf V$ into multiple attention heads, yielding $\mathbf{\hat Q}, \mathbf{\hat K}, \mathbf{\hat V} \in R^{h \times K\times E}$, where $E\times h=C\times \hat H \times \hat W$ and $h$ denotes the number of attention heads. 

Next, standard multi-head and self-attention is applied along the transition state wise. State Transformer can be formulated as follows,
\begin{equation} \label{eq:st}
\begin{gathered}
    \mathbf{Q}=\Phi_Q(S),\mathbf{K}=\Phi_K(S),\mathbf{V}=\Phi_V(S)\\
    Atten= softmax(\mathbf{\hat Q}\cdot \mathbf{\hat K}^T / \sqrt{E})\\ 
    U=\phi_p(\mathbf{\hat V}\cdot Atten) + S\\ 
    T=GDFN(U) + U
\end{gathered}
\end{equation}
where $\Phi(\cdot)$, $\phi(\cdot)$ represent $3\times3$ and $1\times1$ convolutional layers, respectively. $T$ denotes the result of State Attention. 

To better handle complementary information between modalities in Eq.\ref{eq:st}, we utilize the Gated-Dconv Feed-forward Network (GDFN)\cite{zamir2022restormer} proposed in Restormer as the Feed-Forward layer, enabling a gated approach to effectively focus on processing complementary information.

In summary, for the $l$-th CDM module $CDM^{(l)}(\cdot,\cdot)$, we can express its transformation in the following formula,
\begin{equation}
\label{eq:C}
\begin{gathered}
    S^{(l)}=P^{(l)}([V^{(l)};I^{(l)}|c]),T^{(l)}=ST^{(l)}(S^{(l)})\\ 
    Z^{(l)}=[V^{(l)};T^{(l)};I^{(l)}|s] \\ 
    s.t. \quad l \in \{1,2,\cdots, N\}
\end{gathered}
\end{equation}
where $[\cdot|c]$ and $[\cdot|s]$ denote concatenation operators along with the channel wise and the state wise, respectively. The symbols mentioned above are added indexes with $l$ to indicate they belong to the $l$-th CDM. For instance, $ST^{(l)}$ represents the State Transformer in the $l$-th CDM.

\subsection{Decoder Block}
The Decoder Block is used for feature fusion and image reconstruction. Specifically, different layers of the Decoder Block perform feature fusion and image reconstruction at their respective scales. The Decoder Block conducts elementary feature fusion of the the output of CDM through a $3\times 3$ convolutional layer. Subsequently, as illustrated in Fig.\ref{fig:arch} (d), this result is concatenated with the upsampled output of the previous layer of Decoder Block along with the channel wise. Then, through two $3\times3$ convolutional layer and two ReLU activation functions, further fusion and reconstruction at this scale are performed.

In summary, the $l$-th Decoder Block can be expressed as follows,
\begin{equation}
\begin{gathered}
    f^{(N)}=I^{(N)} + V^{(N)}\\
    f^{(l-1)}=DE^{(l)}(f^{(l)},Z^{(l)})\\ 
    s.t. \quad l \in \{1,2,\cdots,N\}
\end{gathered}
\end{equation}
where we perform element-wise addition on the outputs of the $N$-th Encoder Block to obtain $f^{(N)}$, which serves as the input to the $N$-th Decoder Block. The fused image $F$ can be generated through a linear transformation from $f^{(0)}$.

\begin{figure}[tb]
  \centering
  \includegraphics[width=0.4\textwidth,height=0.4\textwidth]{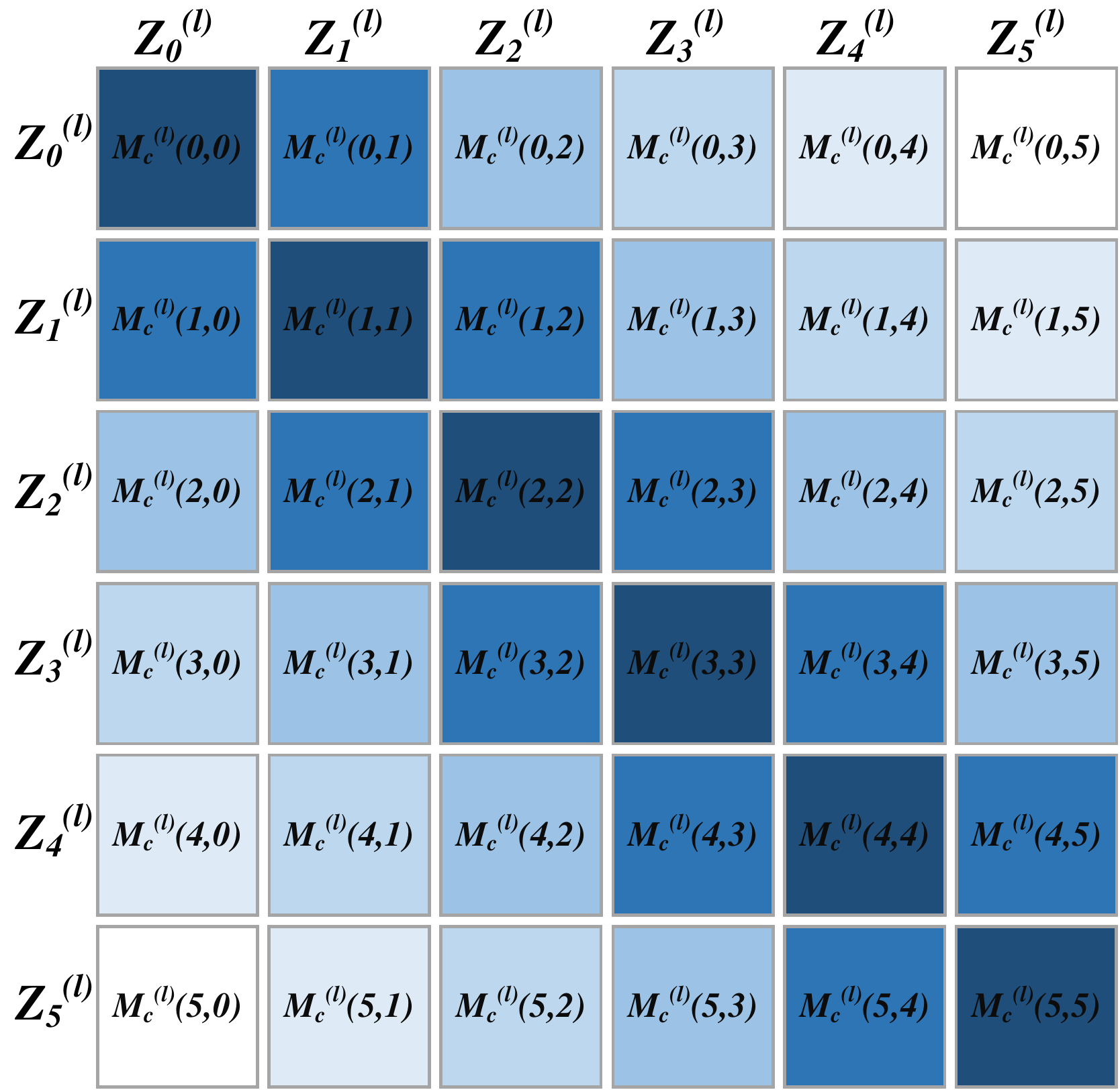}
  \caption{An example of $M^{(l)}_c$ in the $l$-th layer when $K=4$. The color depth represents the constraint of distance, with darker colors indicating them closer to 1.
  }
  \label{loss:fig1}
\end{figure}

\begin{figure}[tb]
  \centering
  \includegraphics[width=0.4\textwidth,height=0.4\textwidth]{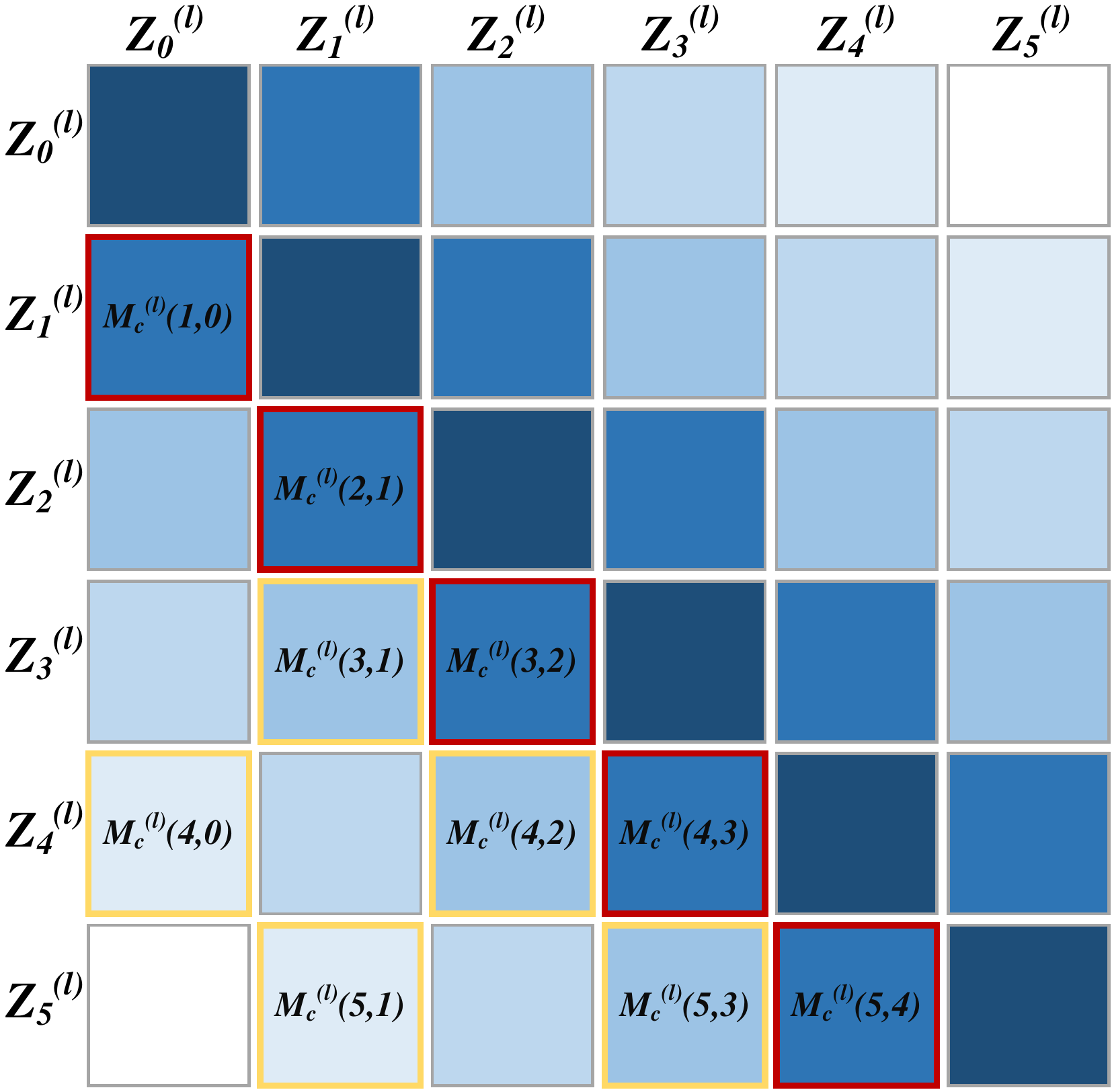}
  \caption{An example of SDS in the $l$-th layer when $K=4$. The yellow boxes represent randomly sampled constraints, while the red boxes represent those calculated consistently each time.
  }
  \label{loss:fig2}
\end{figure}

\begin{figure*}[tb]
  \centering
  \includegraphics[width=1\textwidth,height=0.28\textwidth]{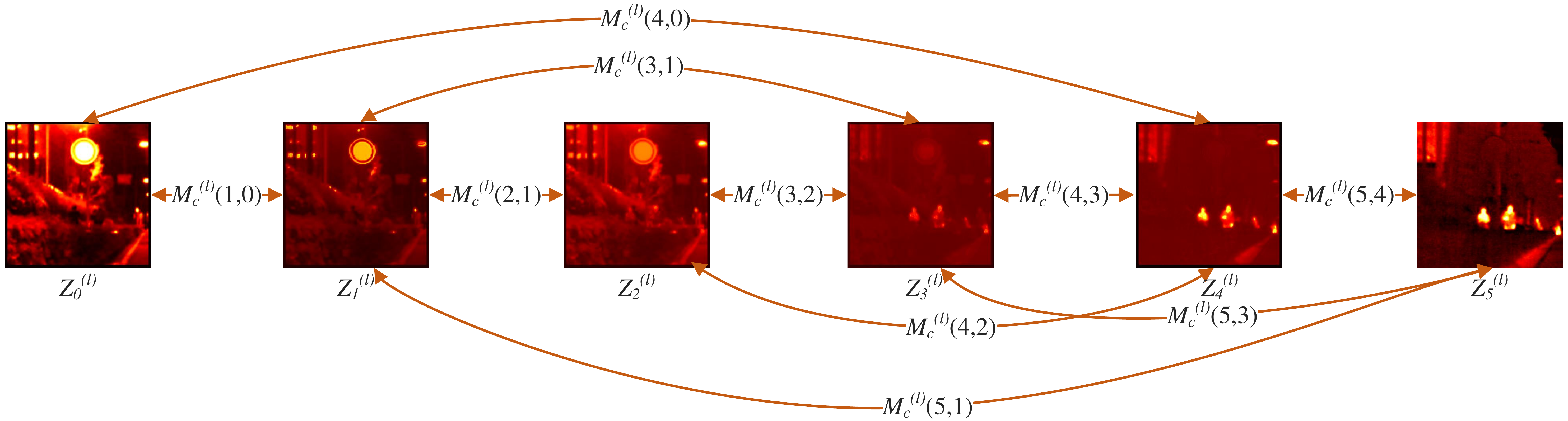}
  \caption{Illustrative example of Fig.\ref{loss:fig2}, where arrows between features represent distance constraints.
  }
  \label{loss:fig3}
\end{figure*}

\subsection{Loss Function}
The loss function $L_{all}$ consists of three parts and can be written as follows:
\begin{gather}
L_{all}=L_{decom}+\alpha_1L_{int}+\alpha_2 L_{grad}
\end{gather}
where $L_{decom}$ is the proposed decomposition loss, $L_{int}=\frac{1}{HW} \Vert F-\max(I,V)\Vert_F^2$ and $L_{grad} =\frac{1}{HW} \Vert |\nabla F|-\max(|\nabla I|,|\nabla V|)\Vert_F^2$ represent the intensity loss and the gradient loss used in SwinFusion \cite{ma2022swinfusion}, respectively\footnote{For more details please refer to SwinFusion \cite{ma2022swinfusion}.}. $\nabla$ denotes the Sobel operator. $\alpha_1$ and $\alpha_2$ are trade-off parameters.

To achieve continuous decomposition, it is necessary to constrain the transition states within a UDFS. Let $\Gamma(\cdot,\cdot)$ denote the distance metric function in the UDFS. For the sake of simplifying subsequent calculations, we set that the greater the distance between two features, the smaller the value computed by $\Gamma$, the formula is given as follows,
\begin{equation}
\begin{gathered}
\Gamma(X, Y) = \frac{1}{C}\sum\limits_{k=1}^C pers(X_k, Y_k) \\
pers(A,B)=  \frac{\sum_{i,j}(A_{i,j} - \bar A)(B_{i,j} - \bar B)}{\sqrt{\sum_{i,j}(A_{i,j} - \bar A)^2}\sqrt{\sum_{i,j}(B_{i,j} - \bar B)^2}}
\end{gathered}
\end{equation}
where, $X, Y \in \mathbb{R}^{C \times H \times W}$ represent two arbitrary features, with $X_k$ and $Y_k$ denoting the corresponding feature maps in the $k$-th channel. Additionally, $\bar{A}$ and $\bar{B}$ represent the mean values of feature maps $A$ and $B$, respectively.

We set the distance between two features with no difference as 1. Initially, we compute the distance between two source features $V^{(l)}$ and $I^{(l)}$ at $l$-th layer in this space using $\Gamma$, denoted as $\mu = \Gamma(V^{(l)}, I^{(l)})$, where $\mu \ll 1$. 

As illustrated in Fig.\ref{loss:fig1}, for the output $Z^{(l)}$ of the $l$-th layer CDM, we calculate the pairwise distances along the direction of state to obtain a symmetric distance matrix $M_{c}$. For the distance matrix $M_c^{(l)}$ of the $l$-th layer:
\begin{equation}
\begin{gathered}
M_c^{(l)}(i,j)=\Gamma(Z^{(l)}_{i},Z^{(l)}_{j})\\ 
s.t. \quad i,j \in \{0,1,\cdots,K+1\}
\end{gathered}
\end{equation}
where $i$ and $j$ are matrix indices, and $Z^{(l)}_{i}$ and $Z^{(l)}_{j}$ represent the $i$-th and $j$-th features at the $l$-th layer. According to Eq.\ref{eq:C}, \(Z_0^{(l)}\) and \(Z_{K+1}^{(l)}\) represent the visible and infrared features at the \(l\)-th layer, respectively. The remaining \(Z_i^{(l)}\) for \(i \in \{1, 2, \cdots, K\}\) correspond to the transitional states. Since the matrix is symmetric, we analyze only its lower triangular part.

By constraining the distance matrix $M_c$, we can impose overall constraints on the decomposition process. In $M_c$, the values on the main diagonal equal to $1$, and we approximate the value at the lower left corner $M_c^{(l)}(K+1, 0)$ to be equal to the distance $\mu$ between the source images. 

For the remaining distances, we let them decay from $1$ to $\mu$ along the direction from the main diagonal to the lower left corner. Thus, we construct a target matrix $M_t$,
\begin{equation}
\begin{gathered}
M_t^{(l)}(i,j) = \Omega(|i-j|,\mu,K+1) \\ 
s.t. \quad i,j \in \{0,1,\cdots,K+1\}
\end{gathered}
\end{equation}
where $\Omega(\cdot,\cdot,\cdot)$ is a designed function used to compute distances during the decay process. 

In this paper, the decay function denoted as Gaussian decay function and the formula are given as follows,
\begin{equation}
\label{eq:decayfun}
\begin{gathered}
\Omega_g(p,\mu,s)=\exp{(-\frac{p^2}{2 * \sigma})}\\ 
s.t. \quad \sigma=-\frac{(s - 1)^2}{2 \ln{(\mu)}}
\end{gathered}
\end{equation}
where $\Omega_g$ denote Gaussian decay which reduces the distance following a Gaussian function along the sub-diagonal direction. We also explore alternative methods for constructing the decay function, with detailed experiments provided in Section \ref{ablation}.

In summary, the decomposition loss $L_{decom}$ can be formulated as follows,
\begin{eqnarray}
\begin{split}
L_{decom} = & \frac{1}{N(K^2 + 3K)}\sum\limits_{l=1}^N\Vert M_c^{(l)} - M_t^{(l)}\Vert_F^2\\
s.t. \quad & M_c^{(l)}(0,K+1)=M_t^{(l)}(0,K+1) = 0 \\ 
& M_c^{(l)}(K+1,0)=M_t^{(l)}(K+1,0) =0
\end{split}
\end{eqnarray}

Note that $M_c^{(l)}(0, K+1)$ and $M_c^{(l)}(K+1,0)$ are not constrained. Because they represent the distances between source features ($V^{(l)}$ and $I^{(l)}$ at $l$-th layer). The constraints within s.t. are established to ensure that the corresponding values at the lower left corner and the upper right corner of the two matrices are equal, thereby relaxing the constraints on the features of the two source images.

Furthermore, the values on the main diagonal are equal. Thus, in both $M_c$ and $M_t$, a total of $(K+2)^2 - (K+2) - 2 = K^2 + 3K$ distances require constraints. In practice, we only compute the lower triangular part of the matrix and ignore the main diagonal and $M_c^{(l)}(K+1,0)$ in $l$-th layer.

\subsection{Support Decomposition Strategy}
The time complexity of the decomposition loss \(L_{decom}\) is \(O(K^2)\) which means that as the number of transition states \(K\) increases, the computational cost of calculating the decomposition loss grows quadratically. 

To enhance the scalability of the proposed fusion model and ensure proper decomposition, we propose the Support Decomposition Strategy (SDS) to compute \(L_{decom}\). Instead of constraining the distances between all features for each decomposition loss calculation, we only constrain a subset. 

Specifically, we constrain the distances between adjacent features and add some randomly sampled distance constraints. For instance, as illustrated in Fig.\ref{loss:fig2} and Fig.\ref{loss:fig3}, when the number of transition states is set to 4, we constrain the distances between all adjacent features (indicated in red) and randomly sample some distances from the remaining ones for constraint (indicated in yellow). Statistically, with sufficient training epochs and ample training data, this partial distance constraint approach approximates the desired decomposition method, thereby reducing the complexity of calculating the decomposition loss.

\begin{figure*}[!tb]
  \centering
  \includegraphics[width=1\textwidth,height=0.37\textwidth]{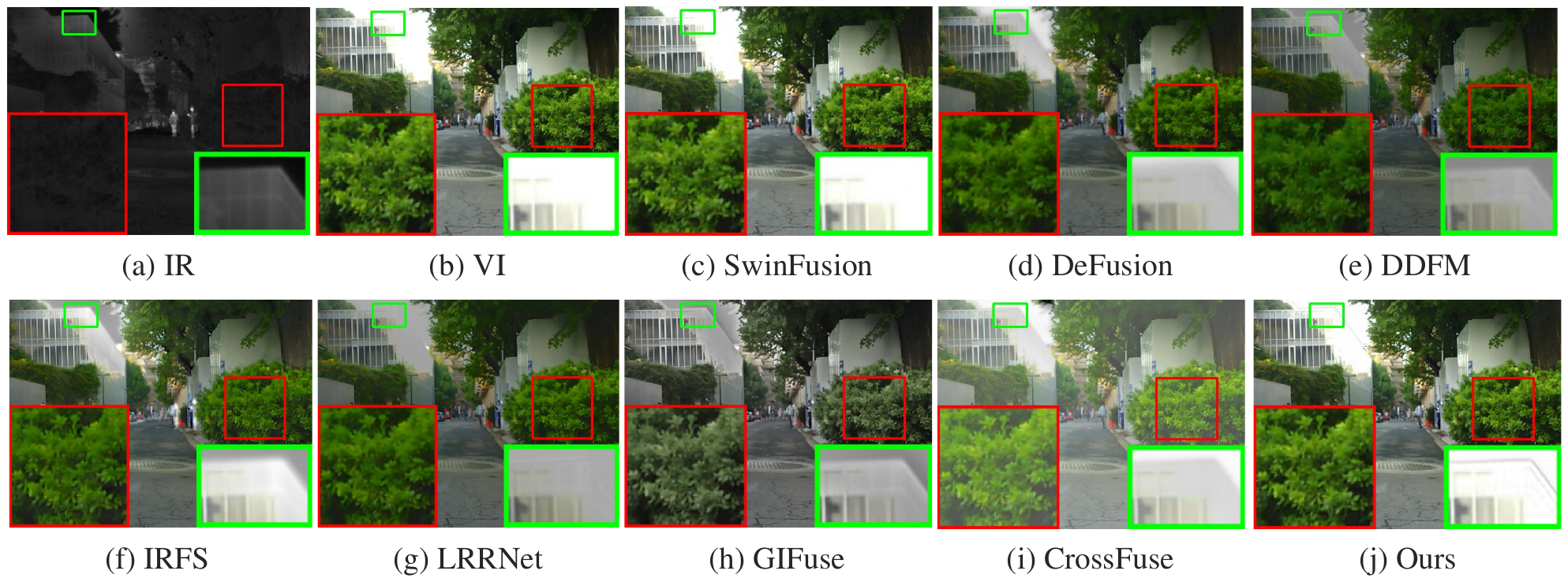}
  \caption{
  Qualitative comparison of image ``00327'' in MSRS.
  }
  \label{fig:MSRS}
\end{figure*}

\begin{table*}[!ht]
\setlength{\tabcolsep}{4mm}{
\caption{Quantitative comparison on MSRS(361 image pairs). The bolded and underlined values represent the best and the second best results.}
\label{tab:cmp_1}
\begin{tabular}{c|c|ccccc|cc}
\hline
   Methods & year & MI             & SF              & AG             & VIF           & Qabf  & LIQE & TOPIQ         \\ \hline
SwinFusion \cite{ma2022swinfusion} & 2022 & \underline{4.529}    & \underline{11.089}    & \underline{3.566}          & \underline{0.99}    & \underline{0.654} & 1.048 & 0.231   \\
DeFusion \cite{liang2022fusion}   & 2022 & 3.345          & 8.606           & 2.781          & 0.763         & 0.530     & 1.025 & 0.214    \\
DDFM \cite{zhao2023ddfm}       & 2023 & 2.728          & 7.388           & 2.522          & 0.743         & 0.474         & 1.025 & 0.217 \\
IRFS \cite{Wang_2023_IF}       & 2023 & 2.151          & 9.888           & 3.155          & 0.735         & 0.477         & 1.020 & 0.224\\
LRRNet \cite{li2023lrr}     & 2023 & 1.108          & 4.162           & 0.966          & 0.194         & 0.103   & 1.036 & 0.229      \\
GIFuse \cite{wang2024general}     & 2024 & 2.409          & 10.425          & 3.310           & 0.857         & 0.625      & \textbf{1.081} & \underline{0.239} \\
CrossFuse \cite{li2024crossfuse}  & 2024 & 3.124          & 9.621           & 3.006          & 0.836         & 0.560       & 1.051 & 0.232 \\ \hline
Conti-Fuse      & Ours & \textbf{5.457} & \textbf{11.478} & \textbf{3.718} & \textbf{1.040} & \textbf{0.710} & \underline{1.067} & \textbf{0.245} \\ \hline
\end{tabular}
}
\end{table*}

For a more general case with \(K\) transition states, at the \(l\)-th layer, we define the sampling strategy as follows,
\begin{eqnarray}
\begin{split}
SDS(\beta,K)=&\{(i+1,i)\}_{i=0}^{K} \cup \\
&\{(u_i,v_i)|(u_i,v_i) \leftarrow \beta\}_{i=0}^{K} \\ 
s.t. \quad & u_i \not = K+1,v_{i}\not = 0,u_i + 1\not = v_i, \\ 
&u_i > v_i, u_i,v_i \in \{0,1,\cdots,K+1\}
\end{split}
\end{eqnarray}
where \(SDS(\cdot,\cdot)\) represents the set of constraints sampled at the \(l\)-th layer, \(\{(u_i, v_i)\}_{i=0}^{K}\) is a set of non-repetitive ordered pairs sampled randomly and \((u_i, v_i)\) denotes the $i$-th sampled unique ordered pair. \(\beta\) is a random seed. Clearly, \(|SDS(\beta, K)| = 2K + 2\). 

Based on the sampled constraints, we can rewrite the decomposition loss as follows,
\begin{eqnarray}
\begin{split}
&L_{decom}= \frac{1}{N(2K+2)} \\ 
&\sum\limits_{l=1}^N \sum\limits_{(i,j) \in SDS(*,K)} [M_c^{(l)}(i,j) - M_t^{(l)}(i,j)]^2
\end{split}
\end{eqnarray}
where \(*\) denotes the random seed determined by the operating system in our training processing, which is used to ensure that the sampling results are as diverse as possible. 

Evidently, by using SDS, we control the time complexity of the decomposition loss to \(O(K)\), thereby significantly enhance the scalability of our proposed model.

\begin{figure*}[!tb]
  \centering
  \includegraphics[width=1\textwidth,height=0.37\textwidth]{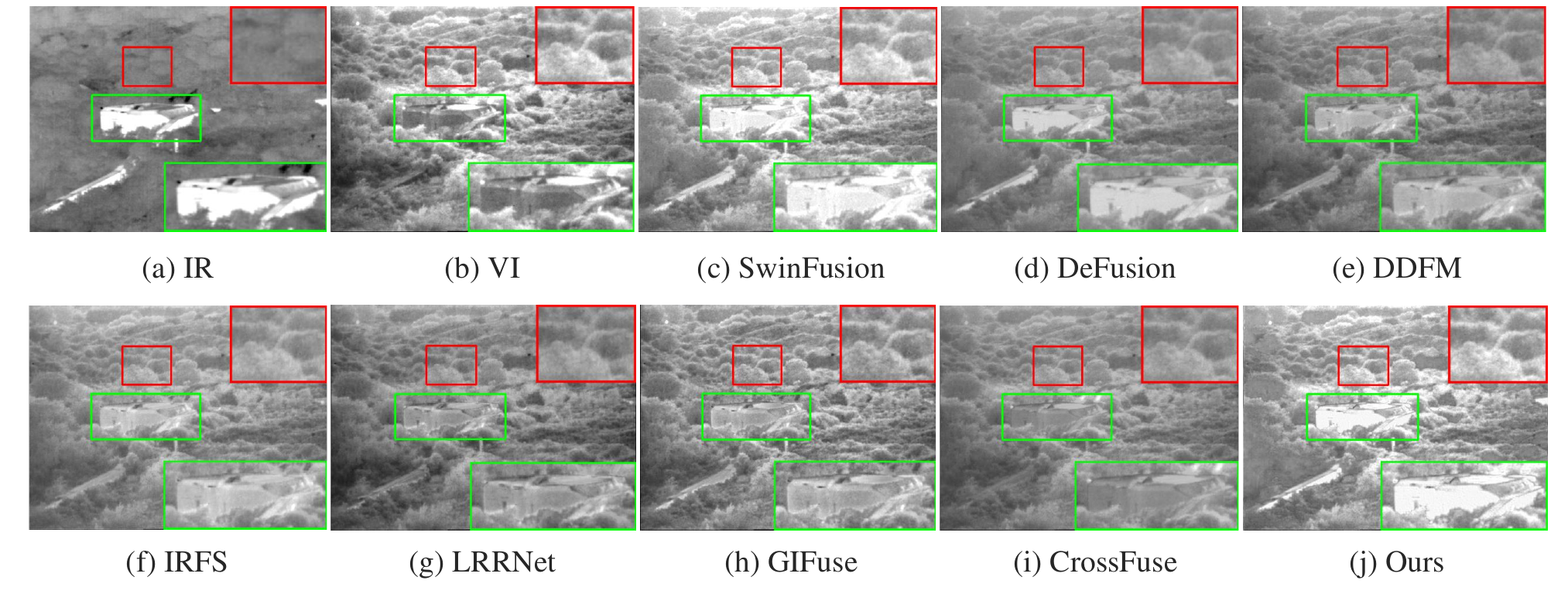}
  \caption{Qualitative comparison of image ``9'' in TNO.
  }
  \label{fig:TNO}
\end{figure*}

\begin{table*}[!ht]
\setlength{\tabcolsep}{4mm}{
\caption{Quantitative comparison on TNO(21 image pairs). The bolded and underlined values represent the best and the second best results.}
\label{tab:cmp_2}
\begin{tabular}{c|c|ccccc|cc}
\hline
  Methods & year & MI             & SF              & AG             & VIF            & Qabf & LIQE & TOPIQ          \\ \hline
SwinFusion \cite{ma2022swinfusion} & 2022 & 3.208          & \underline{10.283}    & \textbf{3.928} & 0.708 & \underline{0.522} & 1.009 & 0.225              \\
DeFusion \cite{liang2022fusion}   & 2022 & 2.530           & 5.652           & 2.298          & 0.509          & 0.332     & 1.013 & 0.213      \\
DDFM \cite{zhao2023ddfm}       & 2023 & 1.605          & 7.852           & 3.183          & 0.258          & 0.227        & 1.012 & 0.201  \\
IRFS \cite{Wang_2023_IF}       & 2023 & 2.094          & 8.347           & 2.950           & 0.563          & 0.416        & 1.013 & 0.223  \\
LRRNet \cite{li2023lrr}     & 2023 & \underline{3.945}    & 6.471           & 2.601          & 0.727          & 0.314    & 1.019 & 0.223      \\
GIFuse \cite{wang2024general}     & 2024 & 1.502          & 9.916           & 3.706          & 0.35           & 0.345     & \underline{1.022} & \underline{0.246}     \\
CrossFuse \cite{li2024crossfuse}  & 2024 & 3.124          & 9.936           & 3.701          & \underline{0.751}    & 0.453  & 1.018 & 0.251        \\ \hline
Conti-Fuse      & Ours & \textbf{4.539} & \textbf{10.479} & \underline{3.773}    & \textbf{0.801} & \textbf{0.553} & \textbf{1.031} & \textbf{0.289} \\ \hline
\end{tabular}
}
\end{table*}

\section{Experiments}
\subsection{Setup}
\subsubsection{Implementation Details}
The number of blocks (Encoder and Decoder) and transition states in our model is set to \(N = 3\) and \(K = 7\). The model width is configured to 8, which corresponds to the number of channels obtained from the linear layer mapping the input source image. Each layer in the CDM contains one State Transformer, and the number of heads in the TSWM is set to 4. We employ average pooling for downsampling and bilinear interpolation for upsampling.
For model training, training images are randomly cropped to \(192 \times 192\), with random flipping being the only data augmentation technique used. The batch size and number of epochs are set to 20 and 250, respectively. To mitigate potential instability during training, we implement gradient clipping to prevent the occurrence of gradient explosion.
AdamW \cite{loshchilov2017decoupled} is utilized as the optimizer, and WarmupCosine serves as the learning rate adjustment strategy. We gradually increase the learning rate from \(10^{-5}\) to \(6 \times 10^{-5}\) during the first 50 epochs, and subsequently, it is gradually decayed to \(5 \times 10^{-6}\) over the remaining epochs. The proposed Gaussian decay function is employed as the decay strategy (Eq. \ref{eq:decayfun}) to compute the decomposition loss, with hyperparameters \(\alpha_1\) and \(\alpha_2\) both set to 15. Our code is implemented using the PyTorch framework, and all experiments are conducted on a NVIDIA GeForce RTX 3090 Ti.

\subsubsection{Datasets and Metrics}

Our model is trained on the training set of MSRS \cite{tang2022piafusion}. For the test set, the testing images on M3FD \cite{liu2022target} and TNO are used in CrossFuse \cite{li2024crossfuse}. The testing images on MSRS \cite{tang2022piafusion} are provided by the dataset.

Furthermore, we evaluate the quality of fused images from both reference and non-reference perspectives using seven objective metrics,
\begin{itemize}
\item Mutual Information (MI) is employed to assess the amount of information retained in the fused image from the two source images. 
\item Spatial Frequency (SF) \cite{fisher1995Image}  is used to evaluate the sharpness of the fused image. 
\item Average Gradient (AG) measures the richness of texture details in the fused image. 
\item Visual Information Fidelity (VIF) \cite{han2013new} quantifies the preservation of visual information between the fused image and the two source images. 
\item Qabf \cite{xydeas2000objective} assesses the representation of salient information in the fused image. 
\item LIQE \cite{zhang2023blind}  employs the image-language model to evaluate image quality, with higher values indicating better quality.
\item TOPIQ\cite{chen2024topiq} utilizes the attention mechanism to assess the levels of distortion and noise in an image, with higher values indicating better quality.
\end{itemize}

\begin{figure*}[!tb]
  \centering
  \includegraphics[width=1\textwidth,height=0.37\textwidth]{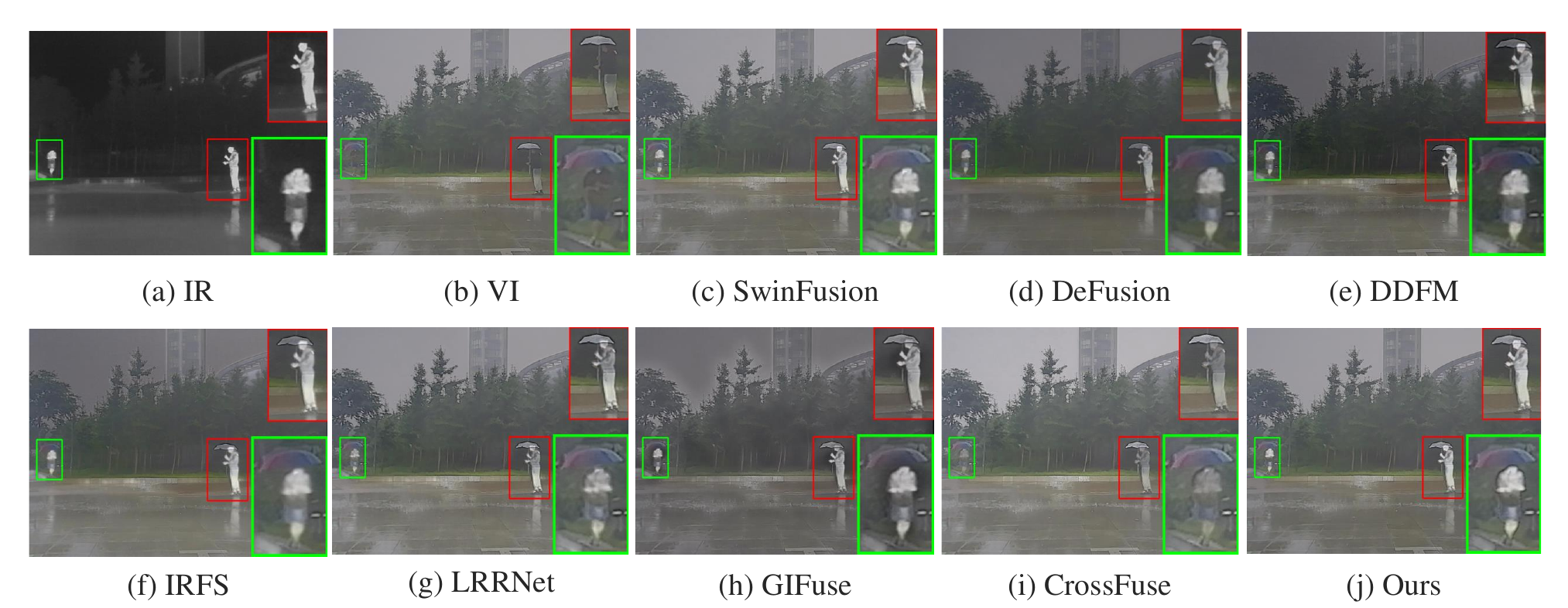}
  \caption{Qualitative comparison of image ``00274'' in M3FD.
  }
  \label{fig:M3FD}
\end{figure*}

\begin{table*}[!ht]
\setlength{\tabcolsep}{4mm}{
\caption{Quantitative comparison on M3FD(300 image pairs). The bolded and underlined values represent the best and the second best results.}
\label{tab:cmp_3}
\begin{tabular}{c|c|ccccc|cc}
\hline
   Methods & year & MI             & SF              & AG             & VIF            & Qabf & LIQE & TOPIQ          \\ \hline
SwinFusion \cite{ma2022swinfusion} & 2022 & 3.907          & 10.869          & 3.363          & 0.897          & 0.621 & 1.399 & 0.363         \\
DeFusion \cite{liang2022fusion}   & 2022 & 2.767          & 6.107           & 1.976          & 0.611          & 0.356  & 1.233 & 0.325       \\
DDFM \cite{zhao2023ddfm}       & 2023 & 2.551          & 7.161           & 2.219          & 0.586          & 0.374  & 1.227 & 0.336        \\
IRFS \cite{Wang_2023_IF}       & 2023 & 2.572          & 8.469           & 2.578          & 0.733          & 0.510 & 1.316 & 0.357         \\
LRRNet \cite{li2023lrr}     & 2023 & \underline{4.801}    & 5.384           & 1.557          & 0.682          & 0.227  & 1.416 & 0.354        \\
GIFuse \cite{wang2024general}     & 2024 & 2.186          & \underline{11.203}    & \underline{3.442}    & \textbf{0.913} & \textbf{0.658} & \underline{1.422} & \textbf{0.385} \\
CrossFuse \cite{li2024crossfuse}  & 2024 & 3.476          & 9.851           & 2.964          & 0.815          & 0.564  & 1.391 & 0.376        \\ \hline
Conti-Fuse      & Ours & \textbf{4.919} & \textbf{11.850} & \textbf{3.475} & \underline{0.902}    & \underline{0.638}  & \textbf{1.423} & \underline{0.377}  \\ \hline
\end{tabular}
}
\end{table*}
Higher values of these five metrics indicate better quality of the fused image.

\subsection{Comparison with Other Methods}

The proposed model is compared with seven state-of-the-art fusion methods, including one diffusion-based method (DDFM \cite{zhao2023ddfm}), two decomposition-based methods (DeFusion \cite{liang2022fusion} and LRRNet \cite{li2023lrr}), one downstream task-integrated method (IRFS \cite{Wang_2023_IF}), one unified fusion method (GIFuse \cite{wang2024general}), and two Transformer-based methods (CrossFuse \cite{li2024crossfuse} and SwinFusion \cite{ma2022swinfusion}). The implementations of these approaches are publicly available.

\subsubsection{Qualitative Comparison}
Fig.\ref{fig:MSRS}, Fig.\ref{fig:TNO}, and Fig.\ref{fig:M3FD} present the qualitative comparison results on MSRS, TNO and M3FD, respectively. It can be observed that our method significantly highlights salient targets and retains more detailed information. 

In Fig.\ref{fig:MSRS}, our method produces more vibrant colors and sharper edges of buildings. In contrast, DDFM, GIFuse, IRFS, DeFusion, and LRRNet exhibit color shifts, CrossFuse results in blurry images, and SwinFusion loses the edge details of buildings. In Fig.\ref{fig:TNO}, our method shows more prominent salient targets (highlighted in the green box) compared to all methods except SwinFusion. 

Additionally, compared to SwinFusion, our method preserves more texture details and provides sharper edges of salient targets. Similarly, in Fig.\ref{fig:M3FD}, our method demonstrates more pronounced salient targets and clearer edges of the figures (green box) compared to other methods. Other methods, such as SwinFusion, DDFM, and IRFS, produce blurrier edges of salient targets. This indicates that our method retains more critical information, demonstrating its effectiveness.


\begin{figure*}[!tb]
  \centering
  \includegraphics[width=1\textwidth,height=0.57\textwidth]{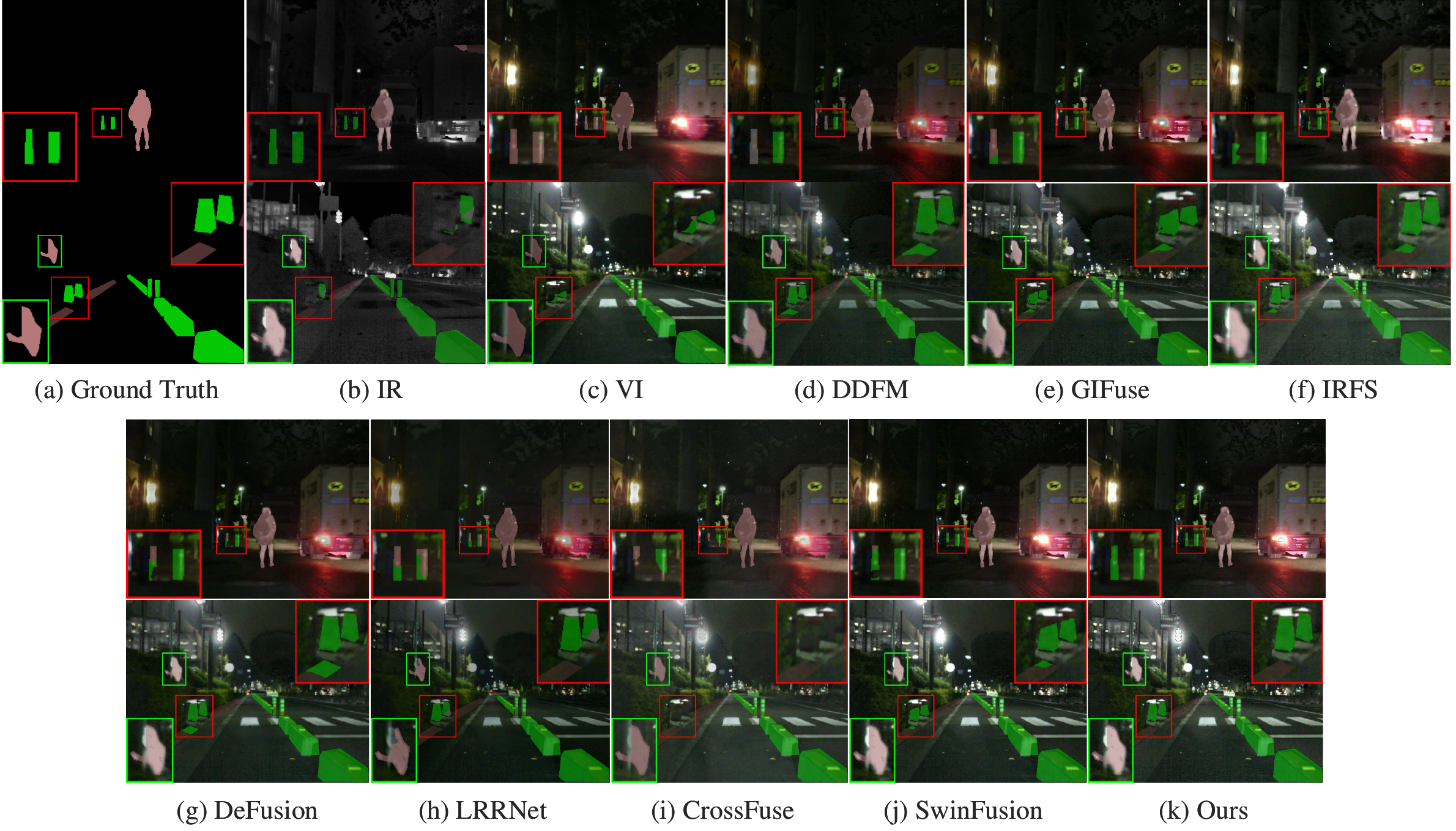}
  \caption{Qualitative evaluation of the multi-modality semantic segmentation task on the MSRS \cite{tang2022piafusion} dataset.
  }
  \label{fig:Seg}
\end{figure*}

\begin{table*}[!ht]
\setlength{\tabcolsep}{1.7mm}{
\caption{Quantitative of the multi-modality semantic segmentation task on MSRS \cite{tang2022piafusion}. The bolded and underlined values represent the best and the second best results.}
\label{tab:seg}
\begin{tabular}{c|cc|cccccccc}
\hline
     & IR    & VI    & DDFM  & GIFuse & IRFS  & DeFusion & LRRNet & CrossFuse   & SwinFusion & Conti-Fuse           \\ \hline
mIoU & 62.58 & 59.78 & 65.96 & 66.78  & 67.00 & 66.59    & 63.99  & \underline{ 67.35} & 65.88      & \textbf{67.80} \\ \hline
\end{tabular}
}
\end{table*}

\subsubsection{Quantitative Comparison}
Table.\ref{tab:cmp_1}, Table.\ref{tab:cmp_2}, and Table.\ref{tab:cmp_3} present the results of the qualitative comparisons on MSRS, TNO and M3FD, respectively. Compared to other methods, our approach consistently achieves superior results across all three datasets. The higher AG and SF values indicate that our method retains more texture information and produces clearer images. 

Additionally, our method demonstrates higher VIF, LIQE and Qabf scores, suggesting that the fused images align better with human visual perception and contain more salient information. Finally, the higher MI score indicates that our method preserves more information from the source images. 

It is noteworthy that our method is trained only on the MSRS dataset and is not fine-tuned on other datasets, which demonstrates its strong generalization capability.

\subsubsection{Multi-Modality Semantic Segmentation}
The multi-modality semantic segmentation task is applied as a high-level visual task to further validate the effectiveness of our approach. Specifically, the fused results from all methods are divided into identical training, testing, and validation sets, and the pretrained B2 model of Segformer \cite{xie2021segformer} is fine-tuned on the fused results of each method. Finally, qualitative and quantitative evaluations are conducted on the respective test sets. 

All methods are fine-tuned using identical experimental parameters, including the same random seed. We utilize AdamW as the optimizer, set the learning rate to \(10^{-5}\), and configure the batch size to 12, without employing any data augmentation techniques. Each method is fine-tuned for 50 epochs.

The multi-modality semantic segmentation experiments are conducted on the MSRS \cite{tang2022piafusion} dataset. The fused results of all methods on the training set are randomly partitioned into new training and validation sets with a 9:1 ratio. The testing set is provided by MSRS \cite{tang2022piafusion}. The mIoU metric is employed for qualitative evaluation of the segmentation results.

As shown in Fig.\ref{fig:Seg}, our method achieves outstanding results in the multimodal object segmentation task. By employing sequential decomposition to obtain multiple intermediate states and thereby mitigating the loss of critical information, our fused results exhibit clearer edges, which leads to superior segmentation performance. 

For instance, our method excels in segmenting the Bump (highlighted in green), a particularly subtle object, outperforming other methods such as SwinFusion, CrossFuse, and DDFM, which struggle with incomplete segmentation due to unclear edges. 

Similarly, as demonstrated in Table.\ref{tab:seg}, our method also demonstrates strong performance in quantitative comparisons. This evidence confirms that our approach effectively retains key information from the source images, benefiting high-level downstream tasks.

\section{Ablation Studies}
In this section, the ablation studies are conducted to verify the rationality of module design, the effectiveness of decomposition loss, and the appropriateness of parameter selection. 

\begin{figure*}[!tb]
  \centering
  \includegraphics[width=1\textwidth,height=0.32\textwidth]{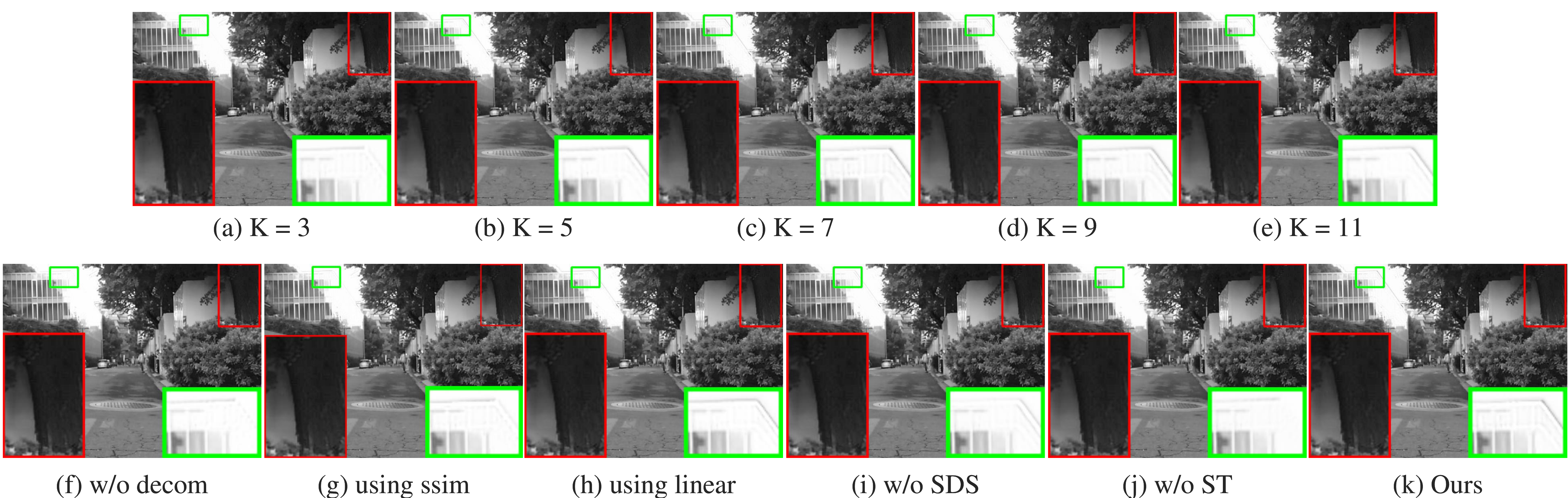}
  \caption{Visualization of ablation experiments in MSRS ``00327D''.
  }
  \label{fig:ablation}
\end{figure*}

\begin{figure*}[!tb]
  \centering
  \includegraphics[width=1\textwidth,height=0.37\textwidth]{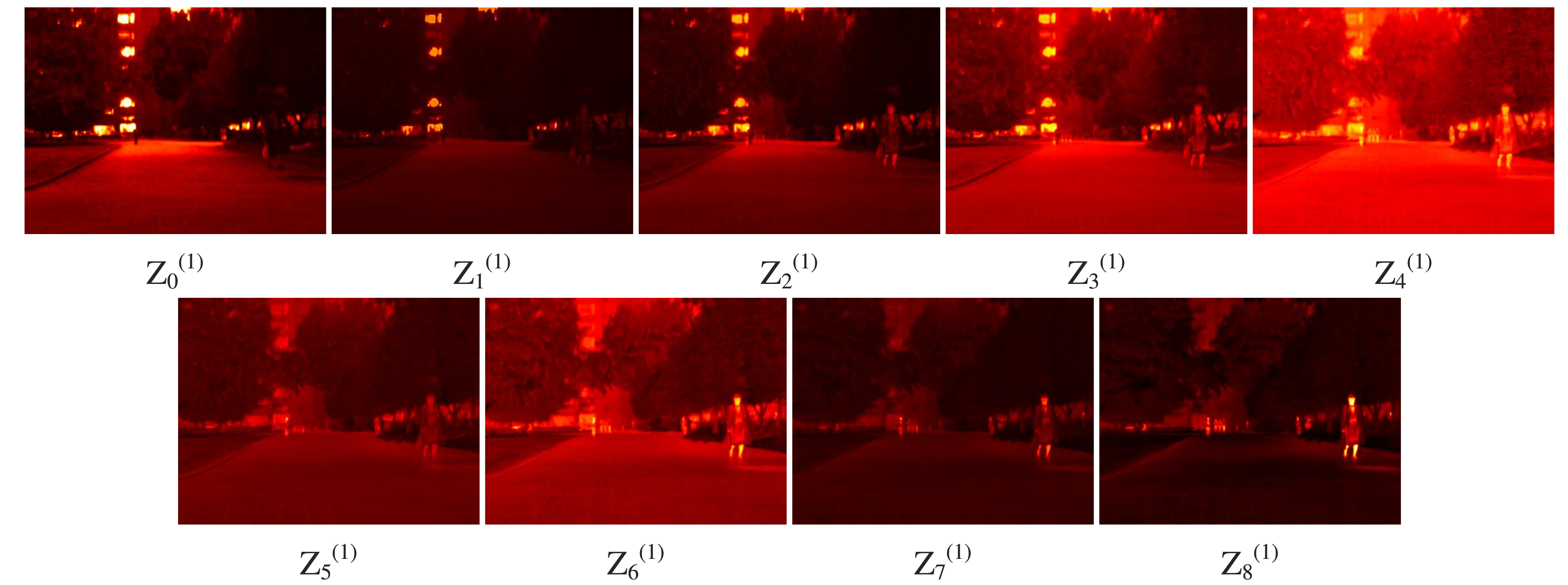}
  \caption{Visualization of transition states and two source images' feature maps in $1$-th layer.
  }
  \label{fig:visual}
\end{figure*}

\subsection{The Number of Transition States}
To determine the most suitable number of transition states \(K\), we conducted ablation experiments. Starting from \(K=3\), we incremented the number of transition states by two in each subsequent experiment. 

As shown in Table.\ref{tab:num_k}, it can be observed that as \(K\) increases, the model's performance gradually improves. However, when \(K\) reaches around 7, the performance gains slow down and show a tendency to decline. Therefore, considering the overall performance of the model, \(K=7\) is reasonable. 

The first row of Fig.\ref{fig:ablation} presents a qualitative comparison for different values of \(K\). It can be seen that when \(K\) is small, the model lacks sufficient representational capacity, leading to significant loss of detail information. Conversely, when \(K\) becomes excessively large and exceeds the suitable representation range, unnecessary redundant information may be overrepresented (for instance, harmful background information from the visible image), while the originally salient information may become diluted by the excess of transition states, leading to the potential loss of some important information.

\begin{table}[!ht]
\tabcolsep=0.07cm
\caption{The ablation experiment of the number of transition states K. The bolded values represent the best results.}
\label{tab:num_k}
\begin{tabular}{c|ccccccc}
\hline
     & MI             & SF              & AG             & VIF           & Qabf   & LIQE & TOPIQ        \\ \hline
K=3  & 5.252          & 11.474          & 3.710           & 1.029         & 0.706 &  1.064 & 0.236       \\
K=5  & 5.374          & 11.464          & 3.682          & 1.039         & 0.708  & 1.062 & 0.238\\
K=7  & \textbf{5.457} & \textbf{11.478} & \textbf{3.718} & \textbf{1.040} & \textbf{0.710} &  \textbf{1.067} &  \textbf{0.245} \\
K=9  & 5.437          & 11.459          & 3.703          & 1.021         & 0.703  & 1.060 & 0.241       \\
K=11 & 5.423          & 11.447          & 3.697          & 1.038         & 0.705  & 1.061 & 0.239         \\ \hline
\end{tabular}
\end{table}

\begin{table*}[!ht]
\tabcolsep=0.3cm
\caption{The ablation experiments of other factors. The bolded values represent the best results.}
\label{tab:others}
\begin{tabular}{c|ccccc|cc|cc}
\hline
           & MI             & SF              & AG             & VIF           & Qabf & LIQE & TOPIQ &  FLOPs$\downarrow$ & Memory$\downarrow$         \\ \hline
w/o decom  & 5.337          & 11.432          & 3.707          & 1.038         & 0.706 & 1.060& 0.238 & - & -         \\
using linear & 5.442          & 11.248          & 3.581          & 1.030          & 0.704 & 1.061& 0.240 & - & -       \\
using ssim     & 5.412          & 11.473          & 3.702         & 1.044         & 0.709 & 1.064& 0.241 & - & -          \\
w/o SDS    & 5.455          & 11.472          & 3.714          & 1.039         & 0.701 & 1.065& 0.244 & 7178M & 19452M         \\
w/o ST     & 5.316          & 11.476          & 3.701          & 1.036         & 0.708& 1.063& 0.239 & - & -          \\
Ours       & \textbf{5.457} & \textbf{11.478} & \textbf{3.718} & \textbf{1.040} & \textbf{0.710} &  \textbf{1.067}& \textbf{0.245} & \textbf{2871M} & \textbf{14356M} \\ \hline
\end{tabular}
\end{table*}

\subsection{Visualization of Transition States}
Fig.\ref{fig:visual} presents the visualization results related to transition states. We visualize the transition state features $\{Z_1^{(1)}, Z_2^{(1)}, \cdots, Z_{7}^{(1)}\}$ of image ``00004N'' from the MSRS dataset, as well as the features of the visible and infrared images $\{Z_0^{(1)}, Z_{8}^{(1)}\}$ according to Eq. \ref{eq:C}. 

From left to right and top to bottom, the feature maps exhibit a continuous changing trend, consistent with our previous discussion. Clearly, these decomposed states are not mutually exclusive. There are overlapping parts and unique parts among them, with each transition state retaining some critical information from the source images. 

For instance, $Z_4^{(1)}$ preserves low-frequency information from both source images, while $Z_5^{(1)}$ retains more high-frequency information from the infrared image. These transition states, which encapsulate rich information from the source images, contribute to retaining more important information from the source images.

\subsection{Other Factors}
\label{ablation}
The ablation experiments on other influencing factors are shown in Table.\ref{tab:others} and Fig.\ref{fig:ablation}. 

\subsubsection{Decomposition loss $L_{decom}$}
Firstly, we conduct an ablation study on the proposed loss function by removing the decomposition loss $L_{decom}$ while keeping other conditions constant. The results, presented in the first row of the table and the ``w/o decom'' plot, indicate that the absence of the decomposition constraint leads to a significant loss of detail in the fused images, resulting in decreased model performance.

\subsubsection{Distance Metric}
We experimented with using another common metric for measuring image distance, SSIM, as a replacement for the Pearson correlation coefficient. The results are shown in the second row of Table \ref{tab:others} and the ``using ssim'' plot in Fig.\ref{fig:ablation}. As can be observed, image quality deteriorates when using SSIM. SSIM emphasizes structural differences within images, whereas CC considers differences based on pixel values. Measuring structural differences between deep features using conventional structural metrics for images is somewhat inappropriate; instead, directly measuring value differences between features in deep representations aligns better with intuition.

\subsubsection{Decay function of $L_{decom}$}
Secondly, we replaced the decay function with a linear one. The linear decay function applies an arithmetic decay with a fixed step size along the sub-diagonal of the distance matrix \(M_c\):
\begin{equation}
\begin{gathered}
\Omega_l(p,\mu,s)=1-p \frac{1-\mu}{s - 1} \\
\end{gathered}
\end{equation}
where \(\Omega_l\) represents the linear decay function. 

The outcomes, shown in the thrid row of the Table.\ref{tab:others} and the ``using linear'' plot, reveal that the window region within the green box becomes blurred, leading to a decline in image quality. We hypothesize that this is due to the linear attenuation being too gradual compared to Gaussian attenuation, failing to effectively separate low-frequency and high-frequency information.

\subsubsection{SDS strategy}
To verify the effectiveness of the SDS strategy, we removed the SDS component. The results, illustrated in the fourth row of the table and the ``w/o SDS'' plot, clearly show a significant increase in the computational load of the loss function, with a quadratic growth trend. 

It can be observed that without SDS, both the computational cost and memory usage during training significantly increase. It should be noted that the computational cost mentioned here only includes the calculation of $L_{decom}$. Additionally, we surprisingly found that the SDS can enhance the model's ability to preserve edge textures. This may be because the randomness introduced by SDS increases the model's robustness.

\subsubsection{State Transformer}
Finally, to validate the effectiveness of the State Transformer, we replaced it with several convolution operations while maintaining a similar number of parameters. 

The results, as shown in the fifth row of the table and the ``w/o ST'' plot, demonstrate that without the State Transformer, the model fails to effectively capture complementary information between transitional states, resulting in a substantial loss of critical information.

\section{Conclusion}
In this paper, a novel fusion framework (Conti-Fuse) based on the designed continuous decomposition strategy is proposed. Conti-Fuse densely samples the trajectory in the unified high-dimensional feature space to decompose the source deep features into multiple transition states, thereby mitigating the loss of critical information from the source input. Additionally, the CDM (including the State Transformer) is introduced to leverage its powerful feature interaction capability, capturing complementary information between transition states and enabling continuous decomposition. Finally, to drive the process of continuous decomposition, a novel and efficient loss function $L_{\text{decom}}$ is proposed.

Both qualitative and quantitative comparisons of our proposed method with seven state-of-the-art (SOTA) approaches from the past three years are conducted. In quantitative evaluations, Conti-Fuse achieved best or second-best results across most metrics, demonstrating the superiority of our approach. Visually, our method effectively preserves the salient information of the source images while capturing fine details, thereby mitigating the loss of important information. In the multimodal semantic segmentation task, Conti-Fuse's fusion results retain more edge information from the source images, which enhances segmentation performance compared to the above methods.








\bibliographystyle{elsarticle-num}
\bibliography{egbib}
\end{document}